\documentclass[sigconf]{acmart}
\usepackage{tabularx} 
\usepackage{ragged2e}

\AtBeginDocument{%
  }

\setcopyright{acmlicensed}
\copyrightyear{2026}
\acmYear{2026}
\acmDOI{XXXXXXX.XXXXXXX}
\acmConference[ICAIF 2026]{RL4LLM-Agents}{Nov 14--15,
  2026}{Milan, Italy}
\acmISBN{978-1-4503-XXXX-X/2018/06}

\begin{document}

\title{Expert-Guided Rubrics as Dense Reward Signals for Reinforcement Learning in Non-Verifiable Domains}

%


\author{Ashish Bansal}
\authornote{Corresponding author.}
\email{ashish\_bansal@vanguard.com}
\affiliation{%
  \institution{The Vanguard Group}
  \city{Malvern}
  \state{Pennsylvania}
  \country{USA}
}

\author{Hossam Amer}
\email{hossam\_amer@vanguard.com}
\affiliation{%
  \institution{The Vanguard Group}
  \city{Toronto}
  \state{Ontario}
  \country{Canada}
}

\author{Keerat Kaur Guliani}
\email{keerat\_guliani@vanguard.com}
\affiliation{%
  \institution{The Vanguard Group}
  \city{Toronto}
  \state{Ontario}
  \country{Canada}
}

\author{Parinita Edke}
\email{parinita\_edke@vanguard.com}
\affiliation{%
  \institution{The Vanguard Group}
  \city{Toronto}
  \state{Ontario}
  \country{Canada}
}

\author{Lesley Matheson}
\email{lesley\_matheson@vanguard.com}
\affiliation{%
  \institution{The Vanguard Group}
  \city{Malvern}
  \state{Pennsylvania}
  \country{USA}
}

\author{Leizer Teran}
\email{leizer\_teran@vanguard.com}
\affiliation{%
  \institution{The Vanguard Group}
  \city{Malvern}
  \state{Pennsylvania}
  \country{USA}
}

%
\renewcommand{\shortauthors}{Bansal et al.}

\begin{abstract}
Large language models (LLMs) are increasingly serving as autonomous agents in non-verifiable domains such as financial services, where they must interpret user intent, rewrite queries, and retrieve information from extensive document collections while satisfying domain-specific requirements. Reinforcement learning (RL) offers a natural way to train such agents from the consequences of their own behavior, but most recent RL successes rely on {verifiable} domains, where reward can be computed automatically as a sparse, binary signal. Retrieval in non-verifiable domains affords no such oracle: judging whether a query rewrite is
``good'' requires judgment that only a subject-matter expert (SME) can
supply even if this judgment is scarce, and that judgment is graded, not binary. When applied in Group Relative Policy Optimization
(GRPO), a widely used policy-gradient algorithm, such sparse rewards make gradients vanish whenever all sampled
outputs in a group succeed or fail together, stalling optimization. Motivated by this reward sparsity, we propose a method that turns a small, fixed investment of SME
judgment into a dense reward, combined with retrieval-based metrics
into a single hybrid, GRPO-compatible reward for the non-verifiable domains. Experimental results demonstrate improvements of up to 6\% in Recall@3 and 6.7\% in MRR@3 relative to sparse-reward baselines.

\end{abstract}


\received{01 October 2026}
\received[revised]{01 October 2026}
\received[accepted]{01 October 2026}

\maketitle

\section{Introduction}


LLMs are increasingly deployed as autonomous agents in non-verifiable domains such as financial services, where they interpret user intent, rewrite queries, and retrieve information from large collections of documents under domain-specific constraints \cite{liu2025finr1, anon2025iff, amer2026distributed}. Reinforcement learning (RL) is a default procedure for post-training such agents, allowing them to explore and learn directly from the consequences of their own behavior~\cite{christiano2017preferences, ouyang2022instructgpt, moradi2025continuous, amer2026flop}.

Recent advances in RL have been particularly pronounced in \emph{verifiable} domains such as math proofs, unit tests, and executable code, where a reward function can check the answer automatically and return a sparse, binary signal~\cite{deepseek2025r1, yang2024selfevolving, gunjal2026rubrics}. This leaves open a
large class of tasks for which no such automatic oracle exists.
Retrieval in non-verifiable domains is one such task: whether a query rewrite is ``good'' depends on regulatory nuance, institution-specific
terminology, and document semantics that only a subject-matter expert
(SME) can judge, and that judgment is inherently graded, not
pass/fail. In this non-verifiable, high-expertise setting, RL cannot
rely on the sparse binary rewards that verifiable domains afford, and
SME time is too scarce to supply feedback at the scale RL normally
demands. We argue this calls for \emph{dense} rewards grounded by a
limited set of SME-derived rubrics: signals that capture degrees of
correctness rather than binary pass/fail, allowing RL to learn
effectively from a limited budget of SME feedback.

\begin{figure*}[h]
  \centering
  \includegraphics[width=\linewidth]{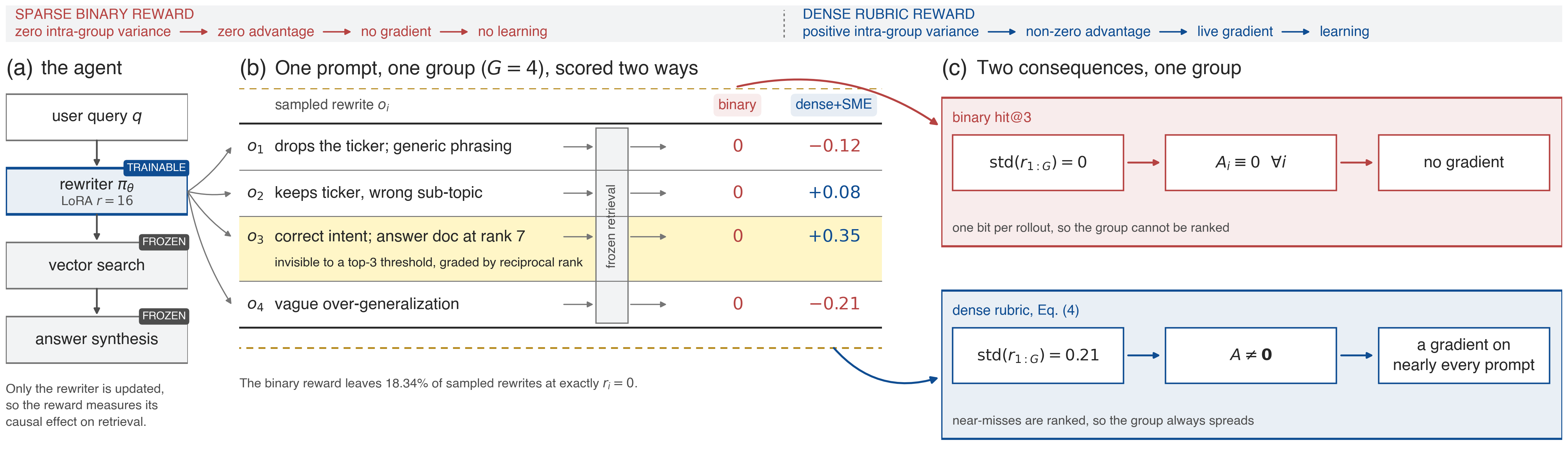}
  \caption{The mechanism at the heart of the paper. \textbf{(a)} The
query-rewriting agent: only the rewriter $\pi_\theta$ is trainable: the
downstream retriever and answer synthesizer are frozen, so the reward
measures the rewriter's causal effect on retrieval. 
\textbf{(b)} One
prompt sampled into a group of $G{=}4$ rewrites (value of $G$ is illustrative here), scored two ways. A
binary top-3 hit reward collapses every rewrite that misses the
threshold to $0$ including $o_3$, which retrieves the answer document
at rank 7 and is invisible to a top-3 cutoff while the dense SME rubric of
Eq.~\eqref{eq:blend} grades the same rewrites continuously and ranks
$o_3$ correctly as the best of the four. \textbf{(c)} The consequence
for GRPO: the binary reward leaves $\mathrm{std}(r_{1:G})=0$ for this
group, so every advantage $A_i$ in Eq.~\eqref{eq:adv} is zero and no
gradient reaches $\pi_\theta$ (the degenerate case of
Eq.~\eqref{eq:degen}); the dense SME reward keeps $\mathrm{std}(r_{1:G})>0$
because near-misses are still ranked, so the group always spreads and
every prompt yields a usable gradient. Dense SME reward is curated using SME judgements [Figure \ref{fig:rubric-loop}, Appendix].}
  \label{fig:main}
\end{figure*}

We ground this argument formally in Group Relative Policy Optimization
(GRPO)~\cite{shao2024deepseekmath}, the RL policy gradient algorithm used throughout
this work. For a group of $G$ completions $\{o_1,\dots,o_G\}$ sampled
per prompt $q$ and scored with rewards $r_i$, GRPO forms the
group-relative advantage
\begin{equation}
\label{eq:adv}
A_i \;=\; \frac{r_i - \operatorname{mean}(r_{1:G})}
                {\operatorname{std}(r_{1:G})}\,.
\end{equation}
Reward \emph{density} is decisive here: under a sparse binary reward,
any prompt whose $G$ samples all succeed or all fail leaves the group
constant, $\operatorname{std}(r_{1:G}) \to 0$, and
Eq.~\eqref{eq:adv} collapses to an uninformative advantage. The policy
is updated only on the fraction of prompts that happen to yield a
mixed group; exploration that moves \emph{closer} to a correct answer
earns no signal, and training dies out (see Figure \ref{fig:main}).

To overcome these challenges, this paper proposes a method that turns a small, fixed investment of SME judgment into a dense, hybrid, GRPO-compatible reward for non-verifiable domains. We \emph{elicit} SME judgment on a small sample of graded rewrites, articulating why one rewrite
retrieves better than another; we \emph{distil} those explanations
into an LLM-executable rubric; and \emph{validate} the rubric
against held-out SME judgments before it is trusted for training. The
validated rubric score is then blended with an objective,
automatically computed retrieval signal into a single dense reward,
which we use to \emph{fine-tune} the rewriter with GRPO. This turns a
fixed, small expert budget into a continuous, scalable reward signal,
and we demonstrate its effectiveness on a retrieval agent. Overall, this paper contributes the following:

\begin{itemize}
\item \textbf{Scalable SME-aligned reward pipeline:} A four-step framework that converts a small, fixed investment of expert judgment into a continuous, dense reward without relying on learned reward models or per-query ground-truth answers.
\item \textbf{Hybrid reward design:} A formulation that bridges reinforcement learning with verifiable rewards (RLVR) and rubric-based RL by combining objective, automated retrieval signals with SME semantic criteria.

\item \textbf{Empirical validation:} Our experiments reveal that reward resolution, not reward correctness alone, is a key determinant of GRPO trainability for query rewriting. The multi-component reward produces a non-zero reward variance in 44\% of groups, compared with only 6.5\% for a binary reward (6.8$\times$), providing a learning signal for substantially more groups and yielding improvements of up to 6\% in Recall@3 and 6.7\% in MRR@3 over sparse-reward baselines.

\end{itemize}

\section{Related Work}

\textbf{RL for LLM alignment} RLHF aligns models to human preference via
a learned reward model optimized with
PPO~\cite{ouyang2022instructgpt,christiano2017preferences,schulman2017ppo};
RLAIF/Constitutional AI replace human labels with model feedback~\cite{bai2022constitutional,lee2023rlaif},
and DPO works with chosen and rejected pairs~\cite{rafailov2023dpo}. GRPO~\cite{shao2024deepseekmath,deepseek2025r1}
discards the critic and estimates advantages from a sampled group. DAPO~\cite{yu2026dapo} resamples zero-variance groups to fix
training degeneracy; Dr.~GRPO~\cite{liu2025understanding}
drops $\operatorname{std}(r_{1:G})$ from Eq.~\eqref{eq:adv} to fix a
related difficulty bias. Unlike GRPO, DAPO, and Dr. GRPO, which modify the optimization process, our work targets reward design for a non-verifiable domain, using GRPO as the underlying policy-gradient algorithm.


\textbf{Reward density and shaping} Sparse terminal rewards are notoriously
hard to learn from; reward shaping provides denser signal while preserving
the optimal policy~\cite{ng1999shaping}. In LLMs, process-supervised and
step-level rewards outperform outcome-only rewards on
reasoning~\cite{lightman2023verify,uesato2022process,setlur2024progress},
and controlled studies show reward granularity interacts with base-model
coverage and prompt distribution to determine
success~\cite{chen2025coverage,clay2026demystify,shao2025spurious,yue2025passk}.
We apply this lens to GRPO, showing density is required not only for
credit assignment but for a non-degenerate advantage in
Eq.~\eqref{eq:adv}.

\textbf{Reward modeling and its failure modes} Learned rewards can be
gamed (reward-model overoptimization, reward hacking, and hackable
process reward
models~\cite{gao2023scaling,skalse2022defining,tiwari2026reward}), 
motivating auditable rewards grounded in explicit criteria.
LLM-as-a-judge~\cite{zheng2023judge} makes it practical to execute a
human-authored rubric at scale, which we exploit to distil SME judgment
into a transparent, weighted reward rather than an opaque scalar model.


\textbf{Financial LLMs and verifiable rewards.} Recent work post-trains
LLMs for finance rather than relying on pretraining alone. Fin-R1~\cite{liu2025finr1}
applies the same SFT-then-GRPO recipe we use, on a 7B model, to
strengthen financial reasoning, showing GRPO is already used for
financial LLM post-training---the exact setting our reward is designed
for. IFF~\cite{anon2025iff} introduces verifiable, chainable constraints
for financial instruction-following and motivates RLVR for finance as
future work, but stops at evaluation, without training a policy against
its verifiers. Our SME rubric plays an analogous role - grounding a
reward in explicit, checkable criteria rather than an opaque judge
score - but closes the loop IFF leaves open by optimizing a policy
against it with GRPO.

\textbf{Retrieval and query rewriting.} Retrieval-augmented
generation~\cite{lewis2020rag} grounds LLM answers in an external
corpus, and learned query rewriting helps when user vocabulary diverges
from corpus terminology~\cite{ma2023rewrite}. RL-trained rewriters fall
into two strands. The first ties reward purely to retrieval outcomes,
with no expert-quality signal: DeepRetrieval~\cite{jiang2025deepretrieval}
and RL-QR~\cite{cha2025annotation} optimize query generation directly
against retrieval metrics via GRPO with no reference queries; ReaLiC \cite{xu2026realic}
adds latent reasoning structure but still no rubric; MaFeRw~\cite{wang2025maferw}
densifies the signal with similarity, ranking, and ROUGE terms but
requires a per-query gold answer. The
second strand - Rubrics-as-Rewards~\cite{gunjal2026rubrics}, \cite{li2026ares},
OpenRubrics~\cite{liu2026openrubrics}, EvoRubric~\cite{guan2026evorubric},
RubricRL~\cite{feng2026rubricrl} - shows that structured, weighted,
LLM-executable rubrics outperform scalar LLM-judge rewards in non-verifiable domains, but only as a reward on the terminal answer and not for query rewrites derived from a small as well as fixed investment of SME feedback merged with retrieval based metrics.


We bridge both strands with a dense reward that blends an objective
retrieval signal with an SME-distilled, validated rubric, applied to the
intermediate query-rewrite action rather than the terminal answer, with
no learned reward model and no per-query gold answer. A small, fixed
investment of SME judgment is distilled once into a weighted rubric and
validated for fidelity, separating and correctly ranking gold vs.\
baseline retrievals before GRPO training, which is needed in retrieval agents operating in non-verifiable domains. 

\section{Methodology}
\label{sec:methodology}

Our work focuses on non-verifiable tasks, where models are post-trained using policy-gradient methods. Throughout our experiments, we use GRPO~\cite{shao2024deepseekmath}, a widely adopted policy optimization method for LLM post-training.

\subsection{Problem Formulation}
\textbf{The GRPO Objective.} For each prompt $q$, GRPO draws $G$ completions from the current policy,
scores them, and updates $\pi_\theta$ with a clipped surrogate objective

\begin{equation}
\label{eq:obj}
\begin{split}
\mathcal{J}(\theta) = \mathbb{E}\!\left[\tfrac{1}{G}\sum_{i=1}^{G}
\tfrac{1}{|o_i|}\sum_{t}
\min\!\big(\rho_{i,t}A_i,\;
\mathrm{clip}(\rho_{i,t},1{-}\varepsilon,1{+}\varepsilon)A_i\big)\right] \\
- \beta\,\mathbb{D}_{\mathrm{KL}}\!\big(\pi_\theta\Vert\pi_{\mathrm{ref}}\big),
\end{split}
\end{equation}
\noindent where $\rho_{i,t}=\pi_\theta(o_{i,t}\!\mid\!q,o_{i,<t})/
\pi_{\theta_{\mathrm{old}}}(o_{i,t}\!\mid\!q,o_{i,<t})$ is the token-level importance ratio and $A_i$ is the group-relative advantage
of Eq.~\eqref{eq:adv}. Two properties of this objective drive the rest
of our argument. First, GRPO carries \emph{no value network}: the group
mean is the only baseline, so the reward's spread \emph{within a group}
is the sole source of learning direction; if that spread vanishes, so
does the gradient - a failure mode we formalize
and repair with our dense reward. Second, $A_i$ is
shared by every token of completion $o_i$; the whole completion is
pushed up or down according to its standing relative to its group. This is precisely why the granularity of the reward
matters and not merely its presence: a reward that ranks completions
correctly relative to one another rather than only classifying each in
isolation-yields a more informative $A_i$, which motivates the
continuously weighted, multi-component reward we construct.


\textbf{Why sparse rewards break GRPO.} Consider a binary reward $r_i \in \{0,1\}$ with per-sample success probability $p$ on a given
prompt. Because each of the $G$ completions is sampled independently from the same policy $\pi_\theta$ conditioned on $q$, the binary outcomes form an i.i.d.\ Bernoulli sample. The group is degenerate (all successes or all failures) with probability
\begin{equation}
\label{eq:degen}
\rho_{\text{deg}} = p^G + (1-p)^G.
\end{equation}


For a degenerate group, $\operatorname{std}(r_{1:G})=0$, so the normalized advantages in Eq.~\eqref{eq:adv} are undefined (and clamped to $0$ in practice), causing the prompt to contribute no gradient. Here $p$ is the policy's per-sample success probability on a given prompt the chance that a single sampled rewrite earns reward $1$ so a low $p$ marks a prompt the policy rarely gets right. Substituting a hard prompt with $p=0.1$ and $G=8$ gives $\rho_{\text{deg}} = 0.1^{8} + 0.9^{8} \approx 0 + 0.43 = 0.43$, dominated by the all-fail term $(1-p)^G$: $43\%$ of such groups produce no learning signal, leaving only $57\%$ with a usable gradient. More clearly, the prompts for which the policy has the lowest success probability are precisely those most likely to yield no gradient signal, making it difficult for training to move $p$ away from the floor. Increasing $G$ reduces this degeneracy, but only by incurring a proportional increase in rollout compute. To gain further insight into Equation \ref{eq:degen}, we empirically measured the fraction of rollout groups at $G=8$ in which all completions receive zero reward under the binary reward scheme. We found that 18.34\% of the rollout groups are zero-reward groups, resulting in no reward variance and consequently impeding training.




A \emph{dense} reward $r_i\in[-1,1]$ that scores \emph{how good} each
completion is, rather than whether it crossed a threshold, almost
always assigns distinct values to the members of a group. Even when all
$G$ rewrites fail to retrieve the answer, one may recover the key entity
while another does not; the dense reward ranks them, $\operatorname{std}$
stays positive, and Eq.~\eqref{eq:adv} produces a gradient that moves the policy toward the marginally better rewrite. Density thus converts
the near-miss structure of the task invisible to a binary reward into
usable signal, which is exactly the regime where credit assignment
under sparse rewards is hardest~\cite{ng1999shaping}.

\textbf{Composite, dense reward} We turn a small, fixed investment of subject-matter-expert (SME)
judgment into a continuous reward through a four-step loop (Figure \ref{fig:rubric-loop}):
\emph{(1) Elicit.} SMEs grade a small sample of rewrite behavior:
roughly $100$ (question, rewrite) pairs sampled (stratified) so the sample spans rewrites that alter vs.\
copy the question and that hit vs.\ miss the answer. For each pair, the expert assigns a headline outcome label ($0/1/2$) indicating whether the documents \emph{retrieved} by the rewrite actually answer the question. The expert also explains \emph{why}, along four recurring axes: whether the rewrite preserved the information need, kept the salient entities, sharpened specificity, and stayed well-formed. These four axes, rather than the individual failure modes, become the rubric criteria.
\emph{(2) Distil.} Those explanations are encoded as an LLM-executable
rubric - a judge prompt with the four named, individually weighted
criteria of Eq.~\eqref{eq:rsme} so the expensive human judgment is
captured once and replayed at scale.
\emph{(3) Validate.} Before training, we check that the rubric is
faithful to the elicited judgment. Using the same stratified set, we
compare the judge's $R_{\text{SME}}$ against the held-out human outcome
labels and verify that it separates strong rewrites from weak
baseline rewrites (verbatim copies and answer-missing queries) and
ranks them in the SME-preferred order, guarding
against a misspecified reward before it can mislead the policy.
\emph{(4) GRPO fine-tune.} We optimize the rewriter against the
validated reward with Eq.~\eqref{eq:obj}, with $R_{\text{SME}}$
computed at each step by the frozen judge via Eq.~\eqref{eq:rsme}.\\

\textbf{Reward composition.} For a question $q$ and a candidate rewrite
$x$, the dense reward, $R$, is a weighted blend of an objective retrieval
signal and an SME rubric signal:
\begin{equation}
\label{eq:blend}
R = a\,R_{\text{retrieval}} + (1-a)\,R_{\text{SME}},
\end{equation}
The two terms are computed differently: $R_{\text{retrieval}}$ is a
deterministic function of $x$ and its retrieved passages, while $R_{\text{SME}}$ is produced by a frozen judge LLM
prompted with a rubric. We describe each in turn.

\emph{(1) Retrieval signal}: $R_{\text{retrieval}}$ is computed by passing $x$ through the frozen retriever, taking the top-$k$ passages, and scoring them with a graded retrieval term $g$ modulated by two multiplicative gates that guard against faithless or malformed rewrites:

\begin{equation}
\label{eq:retrieval}
R_{\text{retrieval}} = g \cdot \gamma_{\text{faith}} \cdot \gamma_{\text{form}}.
\end{equation}

All three terms are computed deterministically from the query, the
rewrite, and the retrieved passages; no model is queried.
\begin{itemize}\itemsep2pt
\item $g \in [0,1]$: a capped DCG@$3$ over the
  answer-bearing passages in the top-$k$ (a passage is answer-bearing if
  it contains a gold answer string),
  \begin{equation}
  \label{eq:driver}
  g = \min\!\Big(1,\ \tfrac{\sum_{r \in H} 1/\log_2(r+1)}
                          {\sum_{i=1}^{3} 1/\log_2(i+1)}\Big),
  \end{equation}
  where $H$ indexes those passages by rank $r$. It rewards both placing an
  answer earlier and covering more answer-bearing passages.
\item $\gamma_{\text{faith}} \in [b,1]$:
  increases with two lexical quantities---the \emph{intent match} (the
  fraction of the question's content words retained in the rewrite) and
  the \emph{entity match} (the fraction of the question's salient
  entities, such as names, numbers, and dates, preserved):
  \begin{equation}
  \gamma_{\text{faith}} = b + (1-b)\,\min\!\big(1,\ \tfrac{0.5\,\mathrm{Intent}+0.5\,\mathrm{Entity}}{0.8}\big).
  \end{equation}

The intent and entity terms are computed using simple lexical string matching overlaps, where we count how many words overlap and obtain a percentage given the input string length.

  
\item $\gamma_{\text{form}} \in [c,1]$:
  penalizes glued or camel-cased tokens, chatter, markup, and multi-line
  output.
  \begin{equation}
  \gamma_{\text{form}} = c + (1-c)\,\min(\mathrm{Wellformed},\mathrm{Conclusive})    
  \end{equation}
  
\end{itemize}

We set $b=0.4$ and $c=0.5$ in all experiments (Table~\ref{tab:reward-weights}). $\gamma_{\text{form}}$ combines two deterministic surface-form scores, each in $[0,1]$, computed directly from the rewrite string with no model call. 
The first, $\mathrm{Wellformed}$, ensures that there are no unnecessary  whitespace-delimited or hyphen tokens. $\mathrm{Wellformed}$ is defined as follows:
\begin{equation}
\label{eq:wellformed}
\mathrm{Wellformed} =
\begin{cases}
1, & m = 0 \quad \text{(no glued tokens)} \\
\max(0,\ 1 - 0.5\,m), & m \geq 1,
\end{cases}
\end{equation}
\noindent where $m$ is the number of glued tokens.

The second, $\mathrm{Conclusive}$, targets string-level artifacts of a query that is not a single clean line. $\mathrm{Conclusive}$ is defined as follows:

\begin{equation}
\label{eq:conclusive}
\mathrm{Conclusive} =
\begin{cases}
1, & p = 0 \quad \text{(clean single-line query)} \\
\max(0,\ 1 - p), & p > 0,
\end{cases}
\end{equation}

where \(p\) is the cumulative penalty obtained by summing the penalties of all artifacts present in the query: \(0.5\) for a multi-line query, \(0.5\) for a conversational preamble, \(0.25\) for a leading quote or bullet, and \(0.25\) for exceeding 240 characters. The score decreases linearly with the accumulated penalty and is clipped at zero through \(\max(0, 1-p)\).




\emph{(2) SME rubric signal}: $R_{\text{SME}}$ is
SME judgment distilled into a weighted rubric and executed by a frozen
judge LLM. The judge is prompted with $q$, the rewrite $x$, and, for each
criterion $c$, a short description with $0/1/2$ anchors (the full prompt
is shown verbatim in Figure~\ref{fig:sme-prompt},
Appendix~\ref{sec:appendix-sme-protocol}); it returns an integer
$s_c \in \{0,1,2\}$ per criterion (as JSON). We normalize by the
top anchor and combine with fixed, SME-elicited weights $w_c$ (summing
to $1$):
\begin{equation}
\label{eq:rsme}
R_{\text{SME}} = \sum_{c \in \mathcal{C}} w_c\,\frac{s_c}{2} \in [0,1].
\end{equation}
The four criteria (weights in Table~\ref{tab:reward-weights}) are intent
fidelity, entity grounding,
specificity/disambiguation, and well-formedness. These
are the \emph{semantic} counterparts of the lexical $\mathrm{Intent}$
and $\mathrm{Entity}$ used in the gates above: the gates are cheap
deterministic floors, while the rubric judges the same properties for
meaning. Keeping the score as a transparent per-criterion sum (rather
than one opaque scalar) makes a low reward traceable to a specific
criterion and, following the Rubrics-as-Rewards
results~\cite{gunjal2026rubrics}, better separates degrees of quality in
non-verifiable domains.

\subsection{Training-stability discipline}

A dense reward removes the gradient-starvation failure but does not by
itself guarantee stable optimization. We add three guardrails.
\textbf{KL anchoring.} A penalty $\beta=0.1$ to a frozen reference
policy $\pi_{\mathrm{ref}}$ (Eq.~\eqref{eq:obj}) keeps the rewriter
from drifting into degenerate text that games the rubric while
abandoning fluency. \textbf{Bounded updates.} Asymmetric PPO-style
clipping ($\varepsilon=0.2$, $\varepsilon_{\mathrm{high}}=0.28$ - a
slightly looser ceiling that permits more upward exploration) together
with gradient-norm clipping caps the size of any single step.
\textbf{Variance monitoring.} Because Eq.~\eqref{eq:adv} is undefined
when group variance vanishes, we track reward variance directly during
training; its collapse is the earliest diagnostic that learning has
stalled (see Section \ref{sec:results}).

\section{Experiment Setup}
\label{sec:exp-setup}

We fine-tune the Qwen3 query-rewriter with GRPO using
\texttt{ms-swift}\allowbreak~\cite{zhao2024swift}, using parameter-efficient LoRA
adapters~\cite{hu2021lora} under a fixed recipe: the same policy,
optimizer, rollout size, and training budget are held constant across
all reward conditions, so any difference in outcome is attributable
to the reward.
All reported experiments use a Qwen3-1.7B rewriter trained with
GRPO over a frozen retriever and a frozen answer synthesizer, on the
Natural Questions split of \textsc{FlashRAG}~\cite{jin2025flashrag}. We split this data set into a train set (79K examples) and an evaluation set (9K examples) for GRPO finetuning. During training, we have $G{=}8$
completions per prompt with vLLM~\cite{kwon2023vllm} colocated
rollout generation. This trains for $1$ epoch over an effective batch of
$64$ rollouts/step ($8$ prompts $\times$ $8$ completions; per-device
batch $2$, gradient accumulation $4$, across $8$ GPUs) under DeepSpeed
ZeRO-2~\cite{rajbhandari2020zero} memory sharding on $8$ $\times$ A100
80\,GB GPUs, and uses the GRPO clipping and KL settings of
Eq.~\eqref{eq:obj}
($\varepsilon{=}0.2$/$\varepsilon_{\text{high}}{=}0.28$,
$\beta{=}0.1$). The reward is registered as an external plugin behind
a common interface, so the dense reward of Eq.~\eqref{eq:blend} and
the sparse binary top-$10$ baseline are swapped in without any other
change to the pipeline. Table~\ref{tab:hyperparams} in Appendix \ref{sec:appendix-config} lists the complete
configuration held constant across all runs. 


\textbf{Reward-granularity ladder.} Rather than a single binary-vs-dense contrast, we compare reward
variants that expose progressively more retrieval signal to the
group. This isolates which form of density
carries the gain, following the treatment of reward granularity as a
first-class axis in recent controlled studies~\cite{clay2026demystify}.
We consider four conditions, reported in Table~\ref{tab:main-results}:
(1) \textbf{Binary hit@$k$}, the sparse top-$k$ baseline; (2) \textbf{DeepRetrieval}~\cite{jiang2025deepretrieval}, a \emph{quantized}
retrieval reward: a step function over recall bands that is finer than a hit/miss reward. We use their Literature Search reward scheme; (3)
\textbf{retrieval-only dense}, the retrieval reward in Equation \ref{eq:retrieval}; and (4) the full blend of Eq.~\eqref{eq:blend}, which adds the SME rubric signal $R_{\text{SME}}$ on top of the retrieval term at retrieval weights $a \in {0.85, 0.70, 0.55}$ (equivalently SME weights $(1-a) \in {0.15, 0.30, 0.45}$). We expect the fraction of non-degenerate groups, and
downstream retrieval quality, to increase with reward density, with
the gap between conditions (1) and (4) as the headline binary-vs-dense
result. The reward weights used throughout our experiments are given in
Table~\ref{tab:reward-weights}. The six sub-score
weights within $R_{\text{retrieval}}$ follow the SME-elicited ordering
of importance described in Section~\ref{sec:methodology} and are
validated (step (3) of our rubric-distillation loop) against the
SME-preferred ranking of gold vs.\ baseline rewrites before training.

\begin{table}[t]
\centering
\caption{Reward composition used in all experiments. $R_{\text{retrieval}}$
is a retrieval driver modulated by two multiplicative guardrail gates with
floors $b$ and $c$ (Eq.~\eqref{eq:retrieval}); $R_{\text{SME}}$ is a weighted
rubric over four SME-elicited criteria (Eq.~\eqref{eq:rsme}). The top-level
blend (Eq.~\eqref{eq:blend}) trades the two off with $a$ (SME weight $1-a$).}
\label{tab:reward-weights}
{\setlength{\tabcolsep}{8pt}
\begin{tabularx}{\linewidth}{>{\RaggedRight\arraybackslash}p{0.26\linewidth}>{\RaggedRight\arraybackslash}Xr}
\toprule
\textbf{Component} & \textbf{Role} & \textbf{Weight} \\
\midrule
\multicolumn{3}{l}{\emph{Top-level blend (Eq.~\eqref{eq:blend})}} \\
$R_{\text{retrieval}}$ & Objective retrieval signal &
  $a \in \{0.85,\ 0.70,\ 0.55\}$ \\
$R_{\text{SME}}$        & Distilled SME rubric score  &
  $1-a \in \{0.15,\ 0.30,\ 0.45\}$ \\
\addlinespace
\multicolumn{3}{l}{\emph{$R_{\text{retrieval}}$: driver $\times$ gates (Eq.~\eqref{eq:retrieval}), multiplicative}} \\
Retrieval driver $g$                        & DCG@$3$-style graded answer-hit score & $[0,1]$ \\
Faithfulness gate $\gamma_{\text{faith}}$   & Lexical intent/entity floor           & $[b,1]$, $b=0.4$ \\
Well-formedness gate $\gamma_{\text{form}}$ & Surface-form floor                    & $[c,1]$, $c=0.5$ \\
\addlinespace
\multicolumn{3}{l}{\emph{$R_{\text{SME}}$: weighted rubric, judge scores $s_c\in\{0,1,2\}$ (Eq.~\eqref{eq:rsme})}} \\
Intent fidelity              & Preserves the information need    & $0.35$ \\
Entity grounding             & Keeps/makes explicit key entities  & $0.25$ \\
Specificity / disambiguation & Adds discriminating terms          & $0.25$ \\
Well-formedness              & Clean, single-line query           & $0.15$ \\
\bottomrule
\end{tabularx}}
\end{table}

\textbf{Reward-degeneracy screen.} We quantify how much learnable signal
each reward exposes \emph{before} committing training compute. For each input prompts from our dataset we sample $G{=}8$ candidate rewrites
from the untrained rewriter, issue one retrieval pass per candidate
against the frozen index, and apply every reward function to those same
fixed top-$10$ passages. We then record $\rho_{\text{deg}}$, the
fraction of prompt groups whose reward vector has
$\operatorname{std}(r_{1:G})\!\approx\!0$ ( Eq.~\eqref{eq:degen}),
and report its complement $1-\rho_{\text{deg}}$ as the \%\,grad column
of Table~\ref{tab:main-results}. Because retrieval is executed once and
every reward scores the same passages, the column isolates the effect of
reward \emph{form} from any difference in retrieval or policy. During training, we additionally log
$\rho_{\text{deg}}$ for every training step.


\textbf{Evaluation and statistical protocol.} The primary metric is retrieval success (answer document in top-$3$ and top-$10$) on
on the evaluation set (9000 examples), with the untrained rewriter (as-is query) as the baseline.  
A
held-out, judge-independent retrieval metric additionally guards against
reward hacking: reward gains that do not transfer to it would indicate
the policy is gaming the rubric rather than improving retrieval.

\section{Results}
\label{sec:results}

We report retrieval at $k\!=\!3$ and $k\!=\!10$
(Recall, MRR) (\S\ref{sec:benchmark}), and separately assess whether retrieval gains survive
scoring by judges that never took part in training (\S\ref{sec:transfer}). We also provide an analysis on whether the rubric matters under retrieval-depth mismatch (typical in production settings) in Appendix \ref{app:crossk}.

\subsection{Dense rubric reward vs.\ sparse RL and prompting}
\label{sec:benchmark}

\begin{table*}[t]
\centering\small
\caption{Retrieval quality on \textsc{FlashRAG}-NQ. \emph{\%\,grad}
is the fraction of GRPO groups with non-zero intra-group reward
variance, i.e.\ $1-\rho_{\text{deg}}$ of Eq.~\eqref{eq:degen}: the
share of prompts that actually produce a gradient this measures the existence and quantity of learnable distinctions
in the reward, not whether the resulting gradient points somewhere
useful. Best per column in bold.}
\label{tab:main-results}
\begin{tabular}{lccccc}
  \toprule
  Method & R@3 & MRR@3 & R@10 & MRR@10 & \%\,grad \\
  \midrule
  \multicolumn{6}{l}{\emph{No RL (prompting only)}}\\
  \quad Raw question, no rewrite   & 71.91 & 0.6296 & 83.32 & 0.6505 & N/A \\
  \quad Few-shot LLM rewrite       & 67.90 & 0.5876 & 80.60 & 0.6107 & N/A \\
  \midrule
  \multicolumn{6}{l}{\emph{Sparse-reward RL}}\\
  \quad Binary hit@$k$         & 67.96 & 0.5885 & 80.82 & 0.6122 & 6.5 \\
  \quad DeepRetrieval (Literature Search)
                                   & 68.54 & 0.5929 & 80.95 & 0.6157 & 17.5 \\
  \midrule
  \multicolumn{6}{l}{\emph{Dense reward (ours)}}\\
  \quad Retrieval-only dense       & 71.94 & 0.6296 & 81.96 & 0.6275 & 41.4 \\
  \quad \;\; $+$ SME rubric, $a\!=\!0.85$  & \textbf{74.00} & 0.6551 & \textbf{83.94} & 0.6690 & \textbf{44.6} \\
  \quad \;\; $+$ SME rubric, $a\!=\!0.70$  & 73.62 & 0.6499 & 83.91 & 0.6686 & 43.9 \\
  \quad \;\; $+$ SME rubric, $a\!=\!0.55$  & 73.92 & \textbf{0.6555} & 83.93 & \textbf{0.6696} & 44.1 \\
  \bottomrule
\end{tabular}
\end{table*}


\textbf{Sparse rewards starve GRPO of gradient.}
The \%\,grad column of Table~\ref{tab:main-results} is the empirical face of
Eq.~\eqref{eq:degen}. A binary hit@$k$ reward leaves only $6.5\%$ of groups with
non-zero reward variance, and DeepRetrieval's reward a single scalar
retrieval metric, denser than a hit/miss bit but still one number per
rollout reaches only $17.5\%$. The retrieval only dense
composite further raises this to $41.4\%$, a $6.4\times$ increase (over binary hit@$k$) in the fraction of prompts that contribute a
usable update. The mechanism argued in \S\ref{sec:methodology} is therefore not
hypothetical: under sparse reward the overwhelming majority of rollout
compute is spent on groups whose advantage is identically zero.

\textbf{Density alone is not sufficient; rubric grounding is what pays.}
In the primary generation setting the retrieval-only dense reward keeps GRPO's gradient alive but does \emph{not} translate
that into retrieval quality: it matches the no-rewrite baseline at
$k\!=\!3$ ($71.94\%$ vs.\ $71.91\%$ R@3, identical MRR@3) and is slightly
\emph{worse} at $k\!=\!10$ ($81.96\%$ vs.\ $83.32\%$ R@10, $0.6275$ vs.\
$0.6505$ MRR@10). Blending in the SME-distilled rubric is what moves the
metrics: the best configuration
reaches $74\%$ R@3 ($+2.09$ over no-rewrite, $+2.06$ over
retrieval-only dense) and $0.6555$ MRR@3 ($+0.026$), with MRR@10 rising
from $0.6505$ to $0.6696$. Read together, these two rows separate the
paper's two claims: reward \emph{density} is what keeps the advantage
non-degenerate, while reward \emph{content} - the expert criteria the
rubric makes gradable is what determines where the surviving gradient
points. A dense reward built from retrieval signals alone satisfies the
first condition and fails the second.

\textbf{The gain is a plateau, not a tuned point.}
Across the three rubric weights, the results are nearly flat: R@3 spans
$73.62$--$74.00$, MRR@3 spans $0.6499$--$0.6555$, and MRR@10 spans
$0.6686$--$0.6696$. Every setting in $a \in \{0.85,\ 0.70,\ 0.55\}$ beats both the
no-rewrite baseline and the retrieval-only dense reward on all four
metrics, and the spread across $a$ ($0.38$ R@3) is roughly five times smaller than the gap to the no-rewrite baseline ($2.09$ R@3). The effect is therefore a property
of adding rubric supervision, not of selecting its weight. Furthermore, the smaller absolute gain at $k\!=\!10$ reflects recall saturation, not
a weaker effect: relative to few-shot prompting the method reduces the miss rate by $19.0\%$ at $k\!=\!3$ and $17.2\%$ at $k\!=\!10$
(gain normalized by the headroom $100-\text{R@}k$), and MRR@10 still
rises $9.6\%$ relative even where recall@10 is near its ceiling.


\subsection{Do the gains transfer, or did we game our own referee?}
\label{sec:transfer}

\begin{table}[t]
\centering\small
\caption{Judge-independent answer quality. AQ is the mean of OpenCompass
and Mistral-7B scores are
rubric-free LLM judges that were never part of the training reward; disagr.\ is their disagreement rate. Adding the SME rubric improves
every column at both $k$. }
\label{tab:judges}
\begin{tabular}{lcccc}
  \toprule
  Run & AQ & OpenCompass & Mistral-7B & disagr. \\
  \midrule
  Retrieval-only, $k$=3         & 0.7432 & 0.6369 & 0.8496 & 0.1199 \\
  \quad $+$SME $a\!=\!0.85$       & 0.7551 & 0.6475 & 0.8628 & 0.1198 \\
  \quad $+$SME $a\!=\!0.70$       & \textbf{0.7554} & 0.6469 & \textbf{0.8639} & 0.1206 \\
  \quad $+$SME $a\!=\!0.55$       & 0.7548 & \textbf{0.6474} & 0.8621 & 0.1198 \\
  \midrule
  Retrieval-only, $k$=10        & 0.7528 & 0.6444 & 0.8603 & 0.1199 \\
  \quad $+$SME $a\!=\!0.85$       & \textbf{0.7616} & 0.6531 & \textbf{0.8683} & 0.1189 \\
  \quad $+$SME $a\!=\!0.70$       & 0.7610 & \textbf{0.6538} & 0.8683 & 0.1203 \\
  \quad $+$SME $a\!=\!0.55$       & 0.7587 & 0.6502 & 0.8671 & 0.1206 \\
  \bottomrule
\end{tabular}
\end{table}

The rubric is the referee during training, so its own score is
guaranteed to rise and is by itself uninformative about whether the
policy improved or merely learned the referee's quirks~\cite{gao2023scaling,
skalse2022defining}. We therefore re-score the
output the synthesized answer with two rubric-free judges from different model families, independent of the reward. The first judge is OpenCompass \cite{cao2026opencompass}, while the second judge is Mistral-7B model \cite{jiang2023mistral}, drawn from a different model family than the SME training judge so that shared biases cannot mask a hack.

For each evaluation question we run the pipeline end to end once. The
trained rewriter's query retrieves the top-$k$ passages, and a
\emph{frozen} synthesizer LLM answers the question from those passages
under a fixed instruction answer using \emph{only} the provided
context, say ``I don't know'' if the answer is absent, and be
concise with the top $k$ passages supplied as numbered context and
greedy decoding. Each judge then scores that answer independently under
a deliberately \emph{rubric-free} prompt (shown verbatim in
Figure~\ref{fig:judge-prompt}, Appendix~\ref{sec:appendix-judge-prompt}):
given the question, the gold answer as a reference, and the synthesized answer, it rates overall
quality from $1$ (poor) to $10$ (excellent) weighing correctness,
completeness, and relevance and returns only the integer, with no
access to our training rubric or checklist. We normalize each score to
$[0,1]$ by dividing by $10$. For a single question, AQ is the mean of
the panel's normalized scores; the AQ column of
Table~\ref{tab:judges} is that mean over all dev queries, and disagreement.\
is the mean per-question standard deviation across the two judges. Both
judges see the identical prompt and decode greedily, so a rise in AQ
reflects answer quality a reward-hacking policy could not manufacture by
replaying the training signal. 


\textbf{Retrieval gains reach the answer.}
Table~\ref{tab:judges} shows the same ordering as
Table~\ref{tab:main-results}. Every rubric-trained run beats its retrieval-only
dense counterpart on Answer Quality (AQ) and on both judges: at $k\!=\!3$, AQ rises
$0.7432\!\rightarrow\!0.7554$ and OpenCompass $0.6369\!\rightarrow\!0.6475$;
at $k\!=\!10$, AQ rises $0.7528\!\rightarrow\!0.7616$ and OpenCompass
$0.6444\!\rightarrow\!0.6538$. The improvement is consistent in sign across all three independent measures, all three rubric weights, and
both retrieval depths. 


\textbf{Two judges, not one signal.}
The two judges disagree on $11.9$--$12.1\%$ of items, and that rate is
stable across all eight runs (spread $0.0017$), so they are scoring
independently rather than collapsing onto a single signal. Their
absolute calibration differs substantially ($\approx\!0.64$ vs.\
$\approx\!0.86$), yet they agree on the coarse ordering of the runs:
both rank retrieval-only dense lowest at each depth and every
rubric-trained run above it. They do not agree on the fine ordering
\emph{within} the three rubric weights ($a\!=\!0.70\!<\!0.55\!<\!0.85$
for OpenCompass, $0.55\!<\!0.85\!<\!0.70$ for Mistral-7B), which is consistent
with the weight sweep being a plateau rather than a ranking.

\textbf{Takeaway.} Algorithm, backbone, data, and compute are held fixed
across conditions; only the reward changes. Sparse rewards leave only 
$\sim\!6.5\%$ of GRPO groups with a meaningful gradient; a dense reward built
from retrieval signals alone fixes that but not the task; adding the
SME-distilled rubric produces the only configuration that improves
retrieval at both depths, and those gains survive three
judge-independent measures. The reward must be both dense enough to keep
the advantage alive and grounded enough to point it somewhere useful.

\section{Conclusion and Future Work}

We presented a framework for training LLM agents in non-verifiable, high-expertise domains, where correctness is graded and expert feedback is scarce. Our approach elicits SME judgment on a small sample of graded rewrites, distils it into an LLM-executable rubric, validates the rubric against held-out SME judgments, and blends its score with an objective retrieval signal into a single dense, GRPO-compatible reward. On a retrieval agent, this reward produced non-zero reward variance in 44.6\% of groups versus 6.5\% under a binary reward (6.8$\times$), and improved Recall@3 by up to 6\% and MRR@3 by up to 6.7\% over sparse-reward baselines. These results suggest that reward resolution, not reward correctness alone, largely determines whether GRPO can learn from a limited expert budget. Although we focused on query rewriting for one dataset, the pipeline makes no domain-specific assumptions beyond the availability of a small set of SME-graded examples, and we expect it to extend to other agentic components and to other domains, such as legal and medical, where judgment is graded and expert time is the bottleneck. In an extension of the proposed method, adaptive group size or difficulty-balanced prompt sampling could allocate rollouts to groups where the dense reward is most informative. This proposed method along with its extensions can be further verified in a wider set of domains and LLM-agents with different SME feedback requirements not only in single turn responses but also in multi-turn responses \cite{jin2025searchr1, kalyan2025longhorizon}.



\section*{Ethics and Privacy Statement}

Our method reduces the expert effort needed to train LLM agents in regulated domains, but it also encodes a small group of SMEs' judgments into a rubric that then acts as the reward at scale, so any bias or blind spot in those judgments can propagate into the trained rewriter and affect which documents users see. In case of non-verifiable tasks such as financial retrieval, they operate over proprietary and potentially sensitive documents and informs consequential decisions, a rewriter tuned to this reward should not be deployed without domain-specific validation, access controls, and human oversight, and rubric agreement with held-out SME judgments should not be taken as evidence of safety or fairness across user groups or institutions. By lowering the cost of training such agents, this work could also make premature deployment easier, and we encourage practitioners to treat the validated rubric as an auditable artifact and to re-validate it when the domain, corpus, or SME panel changes.

\bibliographystyle{ACM-Reference-Format}
\bibliography{custom}

\appendix

\section{Training Configuration}
\label{sec:appendix-config}

Table~\ref{tab:hyperparams} lists the complete GRPO training configuration used for all experiments.

\begin{table}[h]
\centering\small
\caption{Complete GRPO training configuration, held constant across
all reward conditions.
Only the reward function and dynamic resampling differ between the dense
runs and the binary control:
resampling of zero-variance groups is enabled for the dense reward but
disabled for the binary baseline, since resampling would mask the very
gradient stall the control is meant to expose.}
\label{tab:hyperparams}
\begin{tabular}{ll}
  \toprule
  Parameter & Value \\
  \midrule
  \multicolumn{2}{l}{\emph{Model \& adapter}}\\
  \quad Base model              & Qwen3-1.7B \\
  \quad Precision               & bfloat16 \\
  \quad Tuner                   & LoRA \\
  \quad LoRA rank / $\alpha$    & 16 / 32 \\
  \quad Target modules          & all-linear \\
  \addlinespace
  \multicolumn{2}{l}{\emph{GRPO rollout}}\\
  \quad Generations per prompt $G$ & 8 \\
  \quad Max prompt length       & 2048 \\
  \quad Max completion length   & 256 \\
  \quad Sampling temperature    & 1.0 \\
  \quad Top-$p$                 & 0.95 \\

  \addlinespace
  \multicolumn{2}{l}{\emph{Optimization}}\\
  \quad Per-device batch size   & 2 \\
  \quad Gradient accumulation   & 4 \\
  \quad Steps per generation    & 8 \\
  \quad Effective batch         & 64 rollouts/step \\
  \quad Epochs                  & 1 \\
  \quad Learning rate           & $5\times10^{-6}$ \\
  \quad LR schedule             & cosine, $3\%$ warmup \\
  \quad Max grad norm           & 0.5 \\
  \addlinespace
  \multicolumn{2}{l}{\emph{GRPO objective (Eq.~\eqref{eq:obj})}}\\
  \quad KL coefficient $\beta$  & 0.1 \\
  \quad Clip $\varepsilon$ / $\varepsilon_{\mathrm{high}}$ & 0.2 / 0.28 \\
  \quad Importance sampling     & token-level \\
  \addlinespace
  \multicolumn{2}{l}{\emph{Systems}}\\
  \quad Hardware                & $8\times$ A100 80\,GB \\
  \quad Sharding                & DeepSpeed ZeRO-2 \\
  \quad Rollout engine          & vLLM (colocate) \\
  \quad Tensor parallel         & 8 \\
  \quad vLLM GPU mem.\ util.    & 0.6 \\
  \addlinespace
  \multicolumn{2}{l}{\emph{Reward wiring}}\\
  \quad Retrieval depth $k$     & $\{3,\ 10\}$ \\
  \quad SME weight $w$ ($a_2$)  & $\{0,\ 0.15,\ 0.30,\ 0.45\}$ \\
  \quad Regression penalty      & 2.0 \\
  \bottomrule
\end{tabular}
\end{table}

\section{Cross-Depth Transfer of the Reward}
\label{app:crossk}

Both the retrieval term of Eq.~\eqref{eq:blend} and the binary baseline
are computed at a specific retrieval depth $k$: the reward asks whether
the answer document reached the top $k$, and the training signal
therefore inherits that cutoff. The SME rubric terms do not intent
match, entity grounding, and specificity are properties of the rewritten
query, evaluated without reference to any retrieval depth. This suggests
a test of what each reward actually taught the policy. If a rewriter has
learned to satisfy one particular cutoff, moving the evaluation to a
different depth should cost it; if it has learned to write better
queries, it should not.

We therefore train each reward variant at $k\!\in\!\{3,10\}$ and evaluate
every resulting policy at both depths. We call a cell \emph{matched} when
the training and evaluation depths agree (these are the numbers reported
in Table~\ref{tab:crossk}) and \emph{cross} when the policy was trained at
the other depth. Everything else backbone, retriever, synthesizer,
recipe, and held-out query set is held fixed.\\

\begin{table*}[h]
\centering\small
\caption{Cross-depth transfer. \emph{matched} is train-$k$ $=$ eval-$k$;
\emph{cross} is the same reward trained at the other depth. $\Delta$ is
cross $-$ matched, so negative means the depth mismatch cost
performance. The no-rewrite baseline involves no training and is
depth-agnostic. Single run per cell.}
\label{tab:crossk}
\begin{tabular}{lrrr@{\hskip 10pt}rrr}
  \toprule
  & \multicolumn{3}{c}{Evaluated@3} & \multicolumn{3}{c}{Evaluated@10}\\
  \cmidrule(r){2-4}\cmidrule{5-7}
  Reward & matched (train@3) & cross (train@10) & $\Delta$ & matched (train@10) & cross (train@3) & $\Delta$\\
  \midrule
  \multicolumn{7}{l}{\emph{Recall (\%)}}\\
  \quad No rewrite      & \multicolumn{2}{c}{71.91} & --
                        & \multicolumn{2}{c}{83.32} & --\\
  \quad Retrieval only  & 71.94 & 68.98 & $-2.96$ & 81.96 & 80.90 & $-1.06$\\
  \quad $+$SME $a\!=\!0.85$ & 74.00 & 73.58 & $-0.42$ & 83.94 & 82.92 & $-1.02$\\
  \quad $+$SME $a\!=\!0.70$ & 73.62 & 73.56 & $\mathbf{-0.06}$ & 83.91 & 83.10 & $\mathbf{-0.81}$\\
  \quad $+$SME $a\!=\!0.55$ & 73.92 & 73.63 & $-0.29$ & 83.93 & 82.86 & $-1.07$\\
  \midrule
  \multicolumn{7}{l}{\emph{MRR}}\\
  \quad No rewrite      & \multicolumn{2}{c}{0.6296} & --
                        & \multicolumn{2}{c}{0.6505} & --\\
  \quad Retrieval only  & 0.6296 & 0.5955 & $-0.0341$ & 0.6275 & 0.6146 & $-0.0129$\\
  \quad $+$SME $a\!=\!0.85$ & 0.6551 & 0.6498 & $-0.0053$ & 0.6690 & 0.6705 & $+0.0015$\\
  \quad $+$SME $a\!=\!0.70$ & 0.6499 & 0.6495 & $\mathbf{-0.0004}$ & 0.6686 & 0.6668 & $-0.0018$\\
  \quad $+$SME $a\!=\!0.55$ & 0.6555 & 0.6504 & $-0.0051$ & 0.6696 & 0.6713 & $+0.0017$\\
  \bottomrule
\end{tabular}
\end{table*}

\textbf{The SME quality term makes the reward robust to a depth mismatch.}
Adding the rubric almost removes the cost of evaluating at a depth other
than the one trained on. A policy trained at $k\!=\!10$ and evaluated at
$k\!=\!3$ loses only $0.06$--$0.42$ recall points once the rubric is
present, against $2.96$ points for the retrieval-only reward. Ranking
quality is preserved as well: for the SME rewards, cross-depth MRR@10 is
if anything marginally \emph{higher} than the matched value (e.g.\
$0.6705$ cross vs.\ $0.6690$ matched at $a\!=\!0.85$). Because the rubric
grades the rewritten query itself, intent, entity grounding,
specificity, wellformedness, rather than whether a passage cleared a
particular cutoff, the rewrite quality it rewards transfers across the
retrieval depth it was trained on.

\textbf{The pure-retrieval reward overfits to its training depth.}
With no SME term, the reward is tied directly to the top-$k$ retrieval signal it saw in training, so it is by far the largest loser under a depth swap: $-2.96$ recall@3 and $-0.0341$ MRR when trained at $k\!=\!10$ and evaluated at $k\!=\!3$. When evaluated at a different depth, it even drops \emph{below} the untrained baseline ($68.98$ vs.\ $71.91$ recall@3, $0.5955$ vs.\ $0.6296$ MRR), so training against the retrieval reward alone hurts, whereas every rubric-blended policy stays above baseline under the same mismatch.

\textbf{At the deeper cutoff, the mismatch cost is small and even.}
Evaluated at $k\!=\!10$, all four rewards lose a comparable
$\sim\!0.8$--$1.1$ recall points, so the depth mismatch is both smaller
and more uniform than at $k\!=\!3$. Ranking quality is essentially
unaffected for the rubric rewards, two of the three are marginally
higher cross than matched while the retrieval-only reward still slips on
MRR. The deeper cut is more robust depth mismatch across every reward.

\section{Answer-Quality Judge Prompt}
\label{sec:appendix-judge-prompt}

Figure~\ref{fig:judge-prompt} gives the verbatim, rubric-free prompt used
to obtain the judge-independent Answer Quality (AQ) reported in
Table~\ref{tab:judges}. Both cross-family judges receive this identical
prompt; the answer being graded is synthesized by a frozen synthesizer
from the top-$5$ retrieved passages.

\begin{figure}[h]
\centering
\fbox{\begin{minipage}{0.94\linewidth}
\small\ttfamily
\textbf{System:} You are grading the quality of an ANSWER to a
question. Rate overall quality from 1 (poor) to 10 (excellent),
weighing correctness, completeness, and relevance. Do NOT use any
external rubric or checklist give your own holistic judgment.
Respond with ONLY the integer score.\\[4pt]
\textbf{User:} Question: \{question\}\\
Reference answer(s): \{gold\}\\
Answer to grade: \{answer\}\\
Score (1--10):
\end{minipage}}
\caption{Rubric-free judge prompt used to obtain Answer Quality (AQ).
Both cross-family judges (OpenCompass, Mistral-7B) receive this same
prompt with greedy decoding and return a single integer, which we
normalize to $[0,1]$; per-question AQ is the panel mean. The synthesized
answer being graded is produced by a frozen synthesizer from the top-$5$
retrieved passages.}
\label{fig:judge-prompt}
\end{figure}

\section{SME Feedback Elicitation and Annotation Protocol}
\label{sec:appendix-sme-protocol}

\textbf{SME annotation protocol.} To elicit and validate expert
judgment we drew a random sample of $100$ (question, rewrite) pairs
from Natural Questions, stratified across rewritten-vs-verbatim and
hit-vs-miss so the sample spans the quality spectrum. For each pair the
annotator saw the question, the rewrite, the gold answer(s), and the
top-$k$ passages the rewrite retrieved, and assigned (i) a headline
retrieval-quality label $\in\{0,1,2\}$ grading the rewrite by whether
its retrieved passages answer the question, and (ii) four rubric
sub-scores $\in\{0,1,2\}$ (intent fidelity, entity grounding,
specificity/disambiguation, well-formedness) explaining \emph{why}. These
$100$ labels form the gold set against which we validate the automated
rubric judge (step~(3)) before it is trusted as the $R_{\text{SME}}$
reward, and provide the few-shot anchors in the judge prompt.

Figure~\ref{fig:sme-prompt} gives the verbatim prompt the frozen judge
receives to produce $R_{\text{SME}}$.

\begin{figure*}[t]
\centering
\fbox{\begin{minipage}{0.97\linewidth}
\footnotesize\ttfamily\RaggedRight
\textbf{System:} You are a strict retrieval-query reviewer. You grade a
REWRITTEN search query against a fixed rubric. Score each criterion as an
integer 0, 1, or 2 using the anchors given. Judge only the rewrite's
quality as a search query for the user's question --do NOT try to answer
the question yourself. Respond with ONLY a minified JSON object mapping
each criterion key to its integer score. No prose, no code fences.\\[4pt]
\textbf{User:} RUBRIC (score each 0--2):\\
- intent\_fidelity (weight 0.35): Preserves the user's true information
need without drifting to a different question or over-narrowing it.
[0=changes/loses the intent, 1=mostly preserved with some drift,
2=fully preserves the intent]\\
- entity\_grounding (weight 0.25): Surfaces the salient entities the
answer depends on (names, dates, places, domain terms) precisely, without
introducing wrong ones. [0=drops or corrupts key entities, 1=keeps some,
2=keeps/makes explicit all key entities correctly]\\
- specificity\_disambiguation (weight 0.25): Adds discriminating terms
that target answer-bearing evidence and resolves ambiguity -- WITHOUT
keyword-stuffing or padding. [0=vague/generic or stuffed with noise,
1=somewhat specific, 2=precisely specific and discriminating]\\
- well\_formedness (weight 0.15): A single, clean, searchable query: no
chatter, meta-text, quotes, markup, or multi-line output.
[0=malformed/has chatter, 1=minor issues, 2=clean single-line query]\\[4pt]
USER QUESTION:\\
\{question\}\\[2pt]
REWRITTEN QUERY:\\
\{rewrite\}\\[2pt]
Return JSON with exactly these keys: ['intent\_fidelity',
'entity\_grounding', 'specificity\_disambiguation', 'well\_formedness']
\end{minipage}}
\caption{Verbatim SME rubric judge prompt used to compute
$R_{\text{SME}}$ (Eq.~\eqref{eq:rsme}). The frozen judge scores each of
the four criteria on a $0/1/2$ scale from the anchors shown and returns a
minified JSON object; we normalize by the top anchor ($2$) and combine
with the weights of Table~\ref{tab:reward-weights}. The criteria block is
the domain-general distillation used for the pilot; for an expert
domain it is swapped for the SME-elicited rubric while the prompt
scaffold and JSON aggregation are unchanged.}
\label{fig:sme-prompt}
\end{figure*}

\begin{figure*}[t]
\centering
\includegraphics[width=\textwidth]{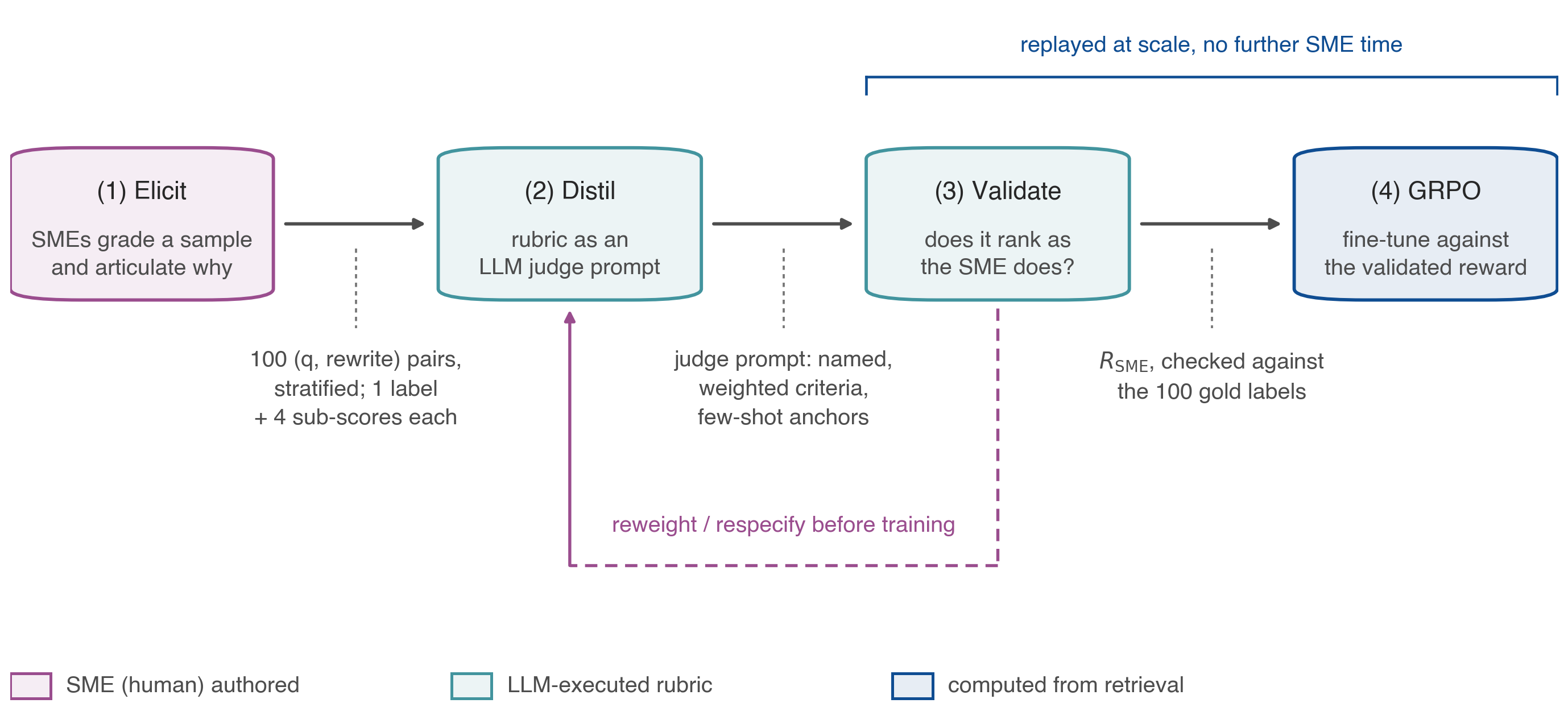}
\Description[Four-stage loop turning SME judgment into a dense reward]{A left-to-right flow of four boxes: Elicit, where SMEs grade a sample and
  articulate why; Distil, where the rubric becomes an LLM judge prompt;
  Validate, which checks that the rubric ranks as the SME does; and GRPO,
  which fine-tunes against the validated reward. Labels beneath the arrows
  name the artifact that passes between stages. A dashed arrow returns from
  Validate to Distil, marked reweight or respecify before training. Box
  colour encodes provenance: SME-authored, LLM-executed, or computed from
  retrieval.}
\caption{From SME judgment to a dense reward. A fixed budget of
  subject-matter-expert grading is elicited once, distilled into an
  LLM-executable rubric of named, individually weighted criteria, and
  validated against the SME-preferred ranking before it is trusted as a
  reward. Only after the rubric separates gold from baseline rewrites do we
  optimize the rewriter against it. Colour marks provenance, separating the
  stages that consume expert time from those that replay that judgment at
  scale.}
\label{fig:rubric-loop}
\end{figure*}

\section{Interpreting the Uplift at $k{=}3$ versus $k{=}10$}
\label{sec:appendix-depth-headroom}
We note first that the absolute uplift at $k{=}10$ over few-shot prompting is not small: our method improves R@10 from 80.60 to 83.94 (+3.34 points, +4.1\% relative) and MRR@10 from 0.6107 to 0.6696 (+9.6\% relative). The apparent difference from the $k{=}3$ gain is explained by headroom: R@10 under few-shot prompting is already 80.60, leaving only 19.4 points of possible improvement, against 32.1 at $k{=}3$. When we normalize each gain by the headroom available at that depth, the improvements are nearly identical---the method reduces the top-3 miss rate by 19.0\% and the top-10 miss rate by 17.2\%. In other words, the smaller absolute gain at $k{=}10$ reflects the saturation of recall at depth 10, not a diminishing effect of the method. This is consistent with the mechanism of query rewriting, which surfaces answer-bearing passages toward the top of the ranking: the effect is largest at low $k$, which is also the operationally decisive regime for RAG, where the synthesizer is conditioned on only a handful of top passages (five in our setup). The rank-sensitive metrics confirm the gain persists at depth---MRR@10 rises 9.6\% relative even where recall@10 is near its ceiling.

\section{Rubric Quality and Robustness to Mis-Specification}
\label{sec:appendix-rubric-quality}

The quality of an SME-distilled rubric is established by validation
rather than assumed. A rubric is admitted to training only after it
separates gold from baseline rewrites in the SME-preferred order and
its score tracks the outcome-grounded human retrieval-quality label on
held-out annotated pairs (\S\ref{sec:methodology}). Two properties make
the reward robust to mis-specification. First, because the rubric is a
transparent, weighted sum of named criteria rather than an opaque
scalar model, a low reward is traceable to a specific criterion and can
be inspected and corrected. Second, the judge-independent, cross-family
answer-quality panel (\S\ref{sec:transfer}) provides an end-to-end
check: a rubric that rewarded spurious surface features would raise the
training reward while leaving these external judges unmoved, whereas we
observe the two move together---the empirical signature of a reward
that captures genuine quality. A residual risk is that a rubric
well-specified on the annotated sample drifts on the broader
population, which periodic SME re-grading is intended to detect.

\end{document}